\documentclass[9pt,journal,compsoc]{IEEEtran}
\usepackage{graphicx}
\usepackage{amsmath}

\ifCLASSOPTIONcompsoc
  \usepackage[nocompress]{cite}
\else
  \usepackage{cite}
\fi
\ifCLASSINFOpdf
\else
\fi
\begin{document}
\title{Unsupervised Representation Learning with Laplacian Pyramid Auto-encoders}

\author{Zhao Qilu
        and Li Zongmin% <-this % stops a space
\IEEEcompsocitemizethanks{\IEEEcompsocthanksitem Zhao Qilu and Li Zongmin are with the Department
of Computer Applications, China University of Petroleum(East China), Huangdao District , Qingdao, China.\protect\\
(E-mail:kg19872006@163.com, 734727745@qq.com)}
% <-this % stops an unwanted space
\thanks{Manuscript received May 1, 2018.}}


\markboth{IEEE TRANSACTIONS ON XXXXX,~Vol.~  , No.~ ,   ~2018}%
{  \MakeLowercase{\textit{et al.}}: Unsupervised Representation Learning with Laplacian Pyramid Auto-encoders}



\IEEEtitleabstractindextext{%
\begin{abstract}
Scale-space representation has been popular in computer vision community due to its theoretical foundation. The motivation for generating a scale-space representation of a given data set originates from the basic observation that real-world objects are composed of different structures at different scales. Hence, it's reasonable to consider learning features with image pyramids generated by smoothing and down-sampling operations. In this paper we propose Laplacian pyramid auto-encoders, a straightforward modification of the deep convolutional auto-encoder architecture, for unsupervised representation learning. The method uses multiple encoding-decoding sub-networks within a Laplacian pyramid framework to reconstruct the original image and the low pass filtered images. The last layer of each encoding sub-network also connects to an encoding layer of the sub-network in the next level, which aims to reverse the process of Laplacian pyramid generation. Experimental results showed that Laplacian pyramid leaded to a more stable and efficient training procedure and improved the performance of the learned representation with scale information.
\end{abstract}

}

\maketitle

\IEEEdisplaynontitleabstractindextext

% \fi
\IEEEpeerreviewmaketitle

\IEEEraisesectionheading{\section{Introduction}\label{sec:introduction}}


\IEEEPARstart{R}{eal} world objects are meaningful only at a certain scale. You might see an apple perfectly on a table. But if looking at the earth, then it simply does not exist. This multi-scale nature of objects is quite common in nature. Scale-space theory is a framework for early visual operations with complementary motivations from physics and biological vision, which has been developed by the computer vision community to handle the multi-scale nature of image data \cite{Lindeberg01}. It is a formal theory for handling visual structures at different scales, by embedding the original image into a one-parameter family of derived images, in which fine-scale structures are successively suppressed. Scale-space representation has a wide application in computer vision. For example, the scale-invariant feature transform (SIFT) \cite{SIFT01}, a successful hand-crafted feature in computer vision to detect and describe local features in images, includes an important stage of key localization, which is defined as minima and maxima of the result of difference of Gaussians (DoG) function applied in scale space to a series of resampled and smoothed images.

In consideration of the successful applications of scale-space representation in hand-crafted feature engineering, it's reasonable to apply it in unsupervised representation learning, especially nowadays when supervised deep learning methods have achieved great success in many tasks, owing to its ability to learn features from raw pixels. Recent work (DeCAF) \cite{DeCAF} has shown that strong generic feature representations can be extracted from the activation of pre-trained networks. DeCAF defined a new visual feature by concatenating the flattened activations of each layer in the pre-trained networks, which is learned on a set of pre-defined object recognition tasks. This feature has shown strong generalization ability when it's applied to new tasks, which suggests that there exists a generically useful feature representation for natural visual data. However, training deep models in a supervised way needs millions of semantically-labeled images which cost lots of manual work. Collecting large labeled datasets is very difficult, and there are diminishing returns of making the dataset larger and larger. Hence, unsupervised representation learning has drawn lots of attention for quick access to arbitrary amounts of data, despite its performance is still limited so far.

The most common method used in unsupervised representation learning is an auto-encoder which learns representations based on an encoder-decoder paradigm. An auto-encoder (AE) \cite{Bourlard1988} is an artificial neural network used for unsupervised learning of efficient coding. It consists of two parts, an encoder which outputs a hidden representation and a decoder which attempts to reconstruct the input from the hidden representation. In this paper we propose Laplacian pyramid auto-encoders (LPAE), a straightforward modification of the deep convolutional auto-encoder architecture, for unsupervised representation learning. The motivation for LPAE originates from a basic observation that real-world objects are composed of different structures at different scales. This implies that real-world objects may appear in different ways depending on the scale of observation. Hence, learning feature representations at multiple scales can make learning system robust to the unknown scale variations that may occur. LPAE is different with the traditional auto-encoder that tries to reconstruct its own inputs. LPAE uses multi-path auto-encoders to reconstruct the Gaussian pyramid from the Laplacian pyramid. Each path has connections with next level, which enables a hierarchical encoding strategy mentioned above.

The rest of the paper is organized as follows. In section2, we discuss related works. Section 3 describes the proposed approach. Evaluation of the proposed approach is presented in section 4.
Finally, conclusions are presented and future research is discussed.

\section{Related Work}

Unsupervised representation learning, aiming to use data without any annotation, is a fairly well studied problem in machine learning community. Examples include dictionary learning \cite{Ng2}, independent component analysis \cite{ICA}, auto-encoders \cite{Bourlard1988}, matrix factorization \cite{Factorization}, and various forms of clustering \cite{kmeans}. We can use K-means algorithm to group an unlabeled data set into k clusters, whose centroids can be used to produce features \cite{Ng1}. Unsupervised dictionary learning exploits the underlying structure of the unlabeled data to optimize dictionary elements. An example of unsupervised dictionary learning is sparse coding, which aims to learn sets of over-complete bases to represent data efficiently \cite{Ng2}.

Recently deep learning methods trained in a supervised way have dramatically improved the state of the art performance on a variety of computer vision tasks. Since supervised deep learning model is capable of learning high-performance visual representations, what about unsupervised deep learning model? Exemplar CNN \cite{ExemplarCNN} proposes a method for training Convolutional Neural Networks(convnet) \cite{LeCun89} through a surrogate task automatically generated from unlabeled images. DCGAN \cite{DCGAN} identified a family of CNN architectures suitable for the adversarial learning framework (GAN) \cite{GAN} which has a wide application in image generation. The most similar work is LAPGAN \cite{LAPGAN}, which uses Laplacian pyramids with convolutional networks in the context of generative model of images.

Another popular method is to train auto-encoders that learns representations based on an encoder-decoder paradigm. Denoising auto-encoders \cite{DenoisingAE} tries to reconstruct the input from a corrupted version of it, which make the hidden layer discover more robust features.
Sparse auto-encoders can learn useful structures in the input data by imposing sparsity on the hidden units during training. Sparsity may be achieved by regularization terms in the loss function \cite{sparseAE}. Contractive auto-encoder \cite{ContractiveAE} adds a regularization term in their loss function that makes the model robust to slight variations of input values. By making strong assumptions concerning the distribution of latent variables, variational auto-encoders \cite{VAE} inherit auto-encoder architecture for learning latent representations. Stacked what-where auto-encoder \cite{SWWE} attempts to learn a factorized representation that encodes invariance and equivariance, and leverage both labeled and unlabeled data to learn this representation in a unified framework. The ladder network \cite{LadderAE} contains several lateral shortcut connections from the encoder to decoder at each level of the hierarchy, and the lateral shortcut connections allow the higher levels of the hierarchy to focus on abstract invariant features.

%\IEEEpubidadjcol

\section{Approach}

The scale-space representation we use is the Laplacian pyramid \cite{LP}. After reviewing this, we introduce our LPAE model which integrates multiple deep convolutional auto-encoders into the framework of a Laplacian pyramid.

\subsection{Laplacian Pyramid}

The Laplacian pyramid is a linear invertible image representation consisting of a set of band-pass images, spaced an octave apart, plus a low-frequency residual. The first step in Laplacian pyramid coding is to low-pass filter the original image $g_0$ to obtain image $g_1$, which is considered a ``reduce'' version of $g_0$ since both resolution and sample density are decreased. In a similar way we form $g_2$ as a reduced version of $g_1$, and so on. Filtering is performed by a procedure equivalent to convolution with one of a family of local, symmetric weighting functions. An important member of this family resembles the Gaussian probability distribution, so the sequence of images $[g_0,g_1,...,g_n]$ is called the Gaussian pyramid. Suppose we have selected the 5-by-5 generating kernel $w$, the level-to-level averaging process is performed by the function REDUCE as below:
\begin{equation} \label{reducefunc}
g_l(i,j)=\sum_{m=-2}^2\sum_{n=-2}^2w(m,n)*g_{l-1}(2i+m,2j+n)
\end{equation}
where $i$ and $j$ denote the coordinate of the pixel.

We define a function EXPAND as the reverse of the function REDUCE. Its effect is to expand an $(M + 1)\times(N + 1)$ image into a $(2M + 1)\times(2N + 1)$ image by interpolating new node values between the given values. Thus, the expand function applied to image $g_l$ of the Gaussian pyramid would yield an image $g_{l}'$ which is the same size as $g_{l-1}$.
\begin{equation} \label{expandfunc}
g_l(i,j)'=4\sum_{m=-2}^2\sum_{n=-2}^2w(m,n)*g_{l}(\frac{i-m}{2},\frac{j-n}{2})
\end{equation}
Only terms for which $\frac{i-m}{2}$ and $\frac{j-n}{2}$ are integers are included in this sum.

The Laplacian pyramid is a sequence of difference images \begin{math}[l_0,l_1,...,l_n]\end{math}. Each is the difference between two levels of the Gaussian pyramid. Thus, for $0<1<n$:
\begin{equation} \label{lp}
l_k=g_k-EXPAND(g_{l+1})
\end{equation}
since there is no image $g_{n+1}$ to serve as the prediction image for $g_n$, we say $l_n=g_n$.

\subsection{Laplacian Pyramid Auto-encoders}
\begin{figure}[!t]
\centering
\includegraphics[width=0.5\textwidth]{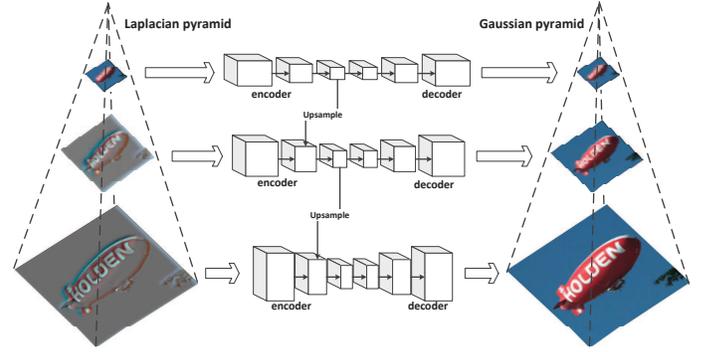}
\caption{The architecture of LPAE. LPAE contains multi-path auto-encoders, which are fed with Laplacian pyramids on the left to reconstruct the corresponding Gaussian pyramids on the right.}
\end{figure}

Suppose we have a Laplacian pyramid \begin{math}[l_0,l_1,...,l_n]\end{math} and the corresponding Gaussian pyramid $[g_0,g_1,...,g_n]$, the aim of our model is to learn a family of hidden representations for the Laplacian pyramid, which can be used to reconstruct the corresponding Gaussian pyramid. A typical architecture of LPAE is shown in Figure 1. We use $E_k()$ and $D_k()$ to denote the encoding network and decoding network at level $k$, separately. The hidden representation $h_k$ is the output of $E_k()$.
\begin{equation} \label{hk}
h_k=\left\{
\begin{split}
E_k(l_k,h_{k+1}),k\neq n \\ E_k(l_k),\quad k=n
\end{split}
\right.
\end{equation}
For each sub-network, the loss function is as below.
\begin{equation} \label{loss}
loss_k=\Vert g_k-D_k(h_k)\Vert_2
\end{equation}
And the total loss is the sum of losses at all levels.
\begin{equation} \label{totalloss}
loss=\sum_{k=0}^nloss_k
\end{equation}

\subsection{Details of the Network Architecture}
\begin{table*}[t]
  \renewcommand{\arraystretch}{1.3}
  \caption{The architecture of various models used in the experiments.}
  \label{LAPGAN}
  \centering
  \begin{tabular}{c|cccc|cccc}
    \hline
    \hline
    4-scales LPAE&&&&convnet&deconvnet&&&\\
    \hline
    \hline
     level 3 &&3*3*64,1&3*3*64,2&3*3*32,1&3*3*64,1&3*3*64,2&3*3*3,1&\\
    \hline
     level 2 &3*3*128,2&3*3*96,1&3*3*96,1&3*3*64,1&3*3*96,1&3*3*96,1&3*3*128,1&3*3*3,2\\
    \hline
     level 1 &5*5*160,2&5*5*128,1&3*3*96,2&3*3*96,1&3*3*96,1&3*3*128,2&5*5*160,1&5*5*3,2\\
    \hline
     level 0 &5*5*192,2&5*5*160,2&3*3*128,2&3*3*128,1&3*3*128,1&3*3*160,2&5*5*192,2&5*5*3,2\\
    \hline
    \hline
    4-scales LAPGAN&&&&convnet&deconvnet&&&\\
    \hline
    \hline
     level 3 $D$&&3*3*64,1&3*3*64,2&3*3*32,1&&&&\\
    \hline
     level 3 $G$&&&&&3*3*64,1&3*3*64,2&3*3*3,1&\\
    \hline
     level 2 $D$&3*3*128,2&3*3*96,1&3*3*96,1&3*3*64,1&&&&\\
    \hline
     level 2 $G$&3*3*128,2&3*3*96,1&3*3*96,1&3*3*64,1&3*3*96,1&3*3*96,1&3*3*128,1&3*3*3,2\\
    \hline
     level 1 $D$&5*5*160,2&5*5*128,1&3*3*96,2&3*3*96,1&&&&\\
    \hline
     level 1 $G$&5*5*160,2&5*5*128,1&3*3*96,2&3*3*96,1&3*3*96,1&3*3*128,2&5*5*160,1&5*5*3,2\\
    \hline
     level 0 $D$&5*5*192,2&5*5*160,2&3*3*128,2&3*3*128,1&&&&\\
    \hline
     level 0 $G$&5*5*192,2&5*5*160,2&3*3*128,2&3*3*128,1&3*3*128,1&3*3*160,2&5*5*192,2&5*5*3,2\\
    \hline
    \hline
    Deep CAE I&&&&convnet&deconvnet&&&\\
    \hline
    \hline
    &5*5*512,2&5*5*512,2&3*3*256,2&3*3*256,1&3*3*256,1&3*3*512,2&3*3*512,2&3*3*3,2\\
    \hline
    \hline
    Deep CAE II&\multicolumn{8}{c}{}\\
    \hline
    \hline
    convnet&5*5*256,1&3*3*128,2&3*3*128,1&\multicolumn{2}{c}{3*3*128,1\;\;\;\quad3*3*128,2}&3*3*96,1&3*3*96,2&3*3*96,1\\
    \hline
    deconvnet&3*3*96,1&3*3*96,2&3*3*128,1&\multicolumn{2}{c}{3*3*128,2\;\;\;\quad3*3*128,1}&3*3*128,1&3*3*256,2&5*5*3,1\\
    \hline
    \hline
  \end{tabular}
\end{table*}

We use a convnet to encode the input, and employ a deconvolutional net (deconvnet) \cite{Deconvnet} to produce the reconstruction at each level. All convolutional layers and deconvolutional layers use ReLU nonlinearity. No fully connected layer has been used, which helps handle input data of different size. Each layer is followed by a batch normalization layer. Batch normalization (BN) layer \cite{BN} is important for the training of deep models based on the CAE, and we give practical proof in the experimental results. We up-sample the outputs of each convnet, and concatenate them with feature maps of a convolutional layer in the next level. This data flow aims to reverse the process of Laplacian pyramid generation.

\section{Experiments}

To compare our approach to other unsupervised feature learning methods, we report classification results on the STL-10 \cite{Ng1}, CIFAR-10 \cite{cifar10} and Caltech-256 \cite{caltech}.

\subsection{Datasets}

STL-10 contains 96x96 pixel images and relatively less labeled data (5,000 training samples, 100,000 unlabeled samples and 8,000 test samples). It is especially well suited for unsupervised learning as it contains a large set of 100,000 unlabeled samples. In all experiments, we trained our model, Deep CAEs and LAPGAN from the unlabeled subset of STL-10. The CIFAR-10 dataset consists of 60,000 32x32 color images in 10 classes, with 6,000 images per class. There are 50,000 training images and 10,000 test images. Since the resolution of CIFAR-10 images is low, we only evaluated 2-scales LPAE and LAPGAN on CIFAR-10. When testing on Catech-256, the images were resized to 96*96 pixels, and we randomly selected 30 samples per class for training and used the rest for testing. For all datasets, we repeated the testing procedure 6 times.

\subsection{Baselines}
In order to evaluate the effectiveness of LPAE, we compared it with the following methods:
\begin{enumerate}
\item Deep convolutional auto-encoders (DCAE): A standard auto-encoder uses a convnet to encode the input, and employs a deconvnet to produce the reconstruction. We use two types of DCAEs with different architectures in the experiments.
\item Laplacian Generative Adversarial Networks (LAPGAN): This method combines the conditional generative adversarial net (CGAN) with a Laplacian pyramid representation to generate natural images in a coarse-to-fine fashion.
\item Exemplar CNN: This method has acchieved the state of the art result for unsupervised learning on several popular datasets, including STL-10, CIFAR-10, Caltech-101 and Caltech-256. Exemplar CNN used several transformations to obtain surrogate data, and train a convnet to learn features that are invariant to these transformations.
\item Other methods for unsupervised representation learning include Convolutional K-means Network (CKN) \cite{Ng3}, Hierarchical Matching Pursuit (HMP) \cite{RGBD} and View-Invariant K-means (VIK) \cite{VIK}.
\end{enumerate}

\subsection{Experimental Setup}
\begin{table*}[t]
  \renewcommand{\arraystretch}{1.3}
  \caption{Classification Performance on Several Datasets (in Percent).}
  \label{classification}
  \centering
  \begin{tabular}{lcccc}
    \hline
    Algorithm&STL10&CIFAR10&Caltech-256(30)&\#feature maps\\
    \hline
    2-scales LPAE&$71.9\pm0.3$&$79.4\pm0.1$&$50.3\pm0.5$&1,088\\
    3-scales LPAE&$73.3\pm0.1$&-&$51.7\pm0.4$&1,472\\
    4-scales LPAE&$72.3\pm0.3$&-&$50.9\pm0.3$&1,632\\
    2-scales LAPGAN&$71.0\pm0.3$&$79.2\pm0.4$&$47.2\pm0.4$&1,088\\
    3-scales LAPGAN&$71.4\pm0.2$&-&$48.7\pm0.4$&1,472\\
    4-scales LAPGAN&$70.5\pm0.2$&-&$47.6\pm0.5$&1,632\\
    Deep CAE I&$70.9\pm0.4$&$76.5\pm0.1$&$45.2\pm0.3$&1,536\\
    Deep CAE II&$67.7\pm0.1$&$73.1\pm0.2$&$41.1\pm0.2$&1,056\\
    \hline
    Convolutional K-means Network \cite{Ng3}&60.1$\pm1$&82.0&-&8,000\\
    Hierarchical Matching Pursuit \cite{RGBD}&64.5$\pm1$&-&-&1,000\\
    View-Invariant K-means \cite{VIK}&63.7&81.9&-&6,400\\
    Exemplar CNN \cite{ExemplarCNN}&74.2$\pm0.4$&84.3&$53.6\pm0.2$&1,884\\
    \hline
    Supervised state of the art&87.26 \cite{cutout}&97.14 \cite{shake}&70.6 \cite{VisCNN}&-\\
    \hline
  \end{tabular}
\end{table*}
\begin{table*}
  \renewcommand{\arraystretch}{1.3}
  \caption{Evaluation of features extracted from each level on STL10 (in Percent). ``C\_$level0$'' means ``discard the features extracted from level 0 and use the rest''.}
  \label{levels}
  \centering
  \begin{tabular}{lccccccccc}
    \hline
    Algorithm&level 0&level 1&level 2&level 3&C\_$level 0$&C\_$level 1$&C\_$level 2$&C\_$level 3$&whole set\\
    \hline
    3-scales LPAE&$69.3\pm0.1$&$66.4\pm0.3$&$58.8\pm0.1$&-&$67.4\pm0.4$&$70.2\pm0.2$&$70.6\pm0.2$&-&$73.3\pm0.1$\\
    4-scales LPAE&$69.5\pm0.2$&$64.0\pm0.2$&$60.6\pm0.4$&$43.3\pm0.3$&$65.8\pm0.4$&$68.7\pm0.2$&$68.8\pm0.2$&$71.8\pm0.1$&$72.3\pm0.3$\\
    \hline
  \end{tabular}
\end{table*}
To make a thorough evaluation of our model, we worked with three network architectures of different scales. We have shown the network architecture of 4-scales LPAE on the top of Table 1. By removing the level 3 of 4-scales LPAE, we get 3-scales LPAE. Likewise, we can get 2-scales LPAE. The architectures of deep CAEs are shown at the bottom of Table 1. It's impossible for LPAE and deep CAE to use same architectures. Thus, there is a question for LAPE and deep CAE, whether different performances are coming from the Lalacian pyramid structure, or simply as a result of more or less parameters. In order to disentangle whether the Laplacian pyramid helps or not, we use two deep CAEs with different number of parameters. The number of parameters of deep CAE I is about 16.7M, and the other has about 2M parameters. The 4-scales LPAE shown in Table 1 has almost 4.3M parameters, which is much less than the deep CAE I. This setup would help us make analysis of the question mentioned above. The architecture of LAPGAN is shown in the middle. By removing the top level and the encoding part of the generator of level 2, we can get a 3-scales LAPGAN. For fair comparison, we make LPAE and LAPGAN share same auto-encoders, and the discriminators LAPGAN use encoding part of the corresponding generators. A softmax layer, omitted in Table 1, follows behind the last convolutional layer of each discriminator, the outputs of which are flattened into vectors to feed in the softmax classifier. Each convolutional (deconvolutional) layer is followed by a batch normalization layer. The numbers in each cell denote the size of receptive field, number of feature maps and stride. For example, ``3*3*64, 1'' means a convolutional (deconvlutional) layer with 3-by-3 receptive field, 64 feature maps and a stride of 1 pixel for each dimension of input. ReLU and BN layers are omitted in the notation.

No pre-processing was applied to training images except ZCA whiting. All models mentioned above were trained with mini-batch Adaptive Moment Estimation (Adam) \cite{Adam} with a mini-batch size of 50. All weights were initialized from a zero-centered Normal distribution with standard deviation 0.02. Learning rate was set to 0.001 in all models. All models were implemented in TensorFlow 1.3 \cite{TensorFlow}.

At test time we applied the discriminators of LAPGAN, the encoding part of deep CAEs and convnets of LPAE as generic feature extractors. To the feature maps of each convolutional layer we applied the max-pooling method that is commonly used for STL-10 and CIFAR-10 dataset. The pooled features were then flattened into vectors, and we concatenated them to form one unique representation of the image. We trained a softmax classifier without regularization on these image representations. For all models, max-pooling results in 16 values per feature map.

\subsection{Classification Performance and Analysis}

We have compared LPAE to several unsupervised feature learning methods, including the current state of the art on each dataset. We also list the state of the art for methods involving supervised feature learning (which is not directly comparable).  In Table 2 we report the classification performances of LPAEs, deep CAEs and LAPGANs, that we have achieved in the experiments. The results of the rest are directly cited from the paper \cite{ExemplarCNN}.

Observations are as follows. First, LPAE and LAPGAN outperformed deep CAEs which didn't consider the scale-space representation. These improvements didn't come from the difference of the number of model parameters, because deep CAE I has more parameters. Second, LPAE performed better than LAPGAN. Training LAPGAN is delicate and unstable. As the discriminator got better, the gradient of the generator vanished. The reasons have been theoretically investigated in \cite{traininggan}. In contrary, training LPAEs is stable, especially with the help of BN layers. Third, LPAE methods didn't achieve the state of the art, but it still outperformed several baselines on STL10 and Caltech-256. Exemplar CNN used various transformations to obtain surrogate data for the CNN training, including scaling the patches by a factor between 0.7 and 1.4. Thus, Exemplar CNN can explore abundant information, which let it learn more discriminative representations. Fourth, both LPAE and LAPGAN performed poor on CIFAR10, which is likely due to the low resolution of CIFAR10 images. Apparently, low resolution fails to provide significant scale-space information. Fifth, the performances of both LPAE and LAPGAN didn't increase with the number of scales. The down-sampled image at level 3 has a very low resolution, which failed to provide discriminative information as shown in Table 3. It's clear that the classification accuracy dropped slightly when discarding features extracted from level 3, and its performance is also pretty poor. The results shown in Table 3 also indicate that coarser levels perform worse than finer ones, but they still contain their own specific information which is useful for image representation. Besides level of scale, it's also related with the number of features that the performance of each level dropped as the scale became coarser.

The dimensions of the learned representations had influences on the classification performance as shown in Table 4. Increasing the representation dimensionality would improve the performance at first, but redundant features became more and more, which made the accuracy dropped slightly. As we can see,  extracting 9 or 16 values per feature map would be appropriate for the models in the experiments.
\begin{table}
  \renewcommand{\arraystretch}{1.3}
  \caption{Evaluation of the influence of representation dimensionality on STL10 (in Percent). ``4/'' means that we extracted 4 values per feature map to form the representation.}
  \label{dimension}
  \centering
  \begin{tabular}{lcccc}
    \hline
    Algorithm&4/&9/&16/&24/\\
    \hline
    2-LPAE&$64.7\pm0.2$&$69.2\pm0.3$&$71.9\pm0.3$&$71.4\pm0.2$\\
    3-LPAE&$69.0\pm0.2$&$72.1\pm0.4$&$73.3\pm0.1$&$72.9\pm0.4$\\
    4-LPAE&$69.3\pm0.3$&$71.5\pm0.4$&$72.3\pm0.3$&$71.7\pm0.1$\\
    2-LAPGAN&$68.4\pm0.6$&$70.7\pm0.4$&$71.0\pm0.3$&$70.5\pm0.1$\\
    3-LAPGAN&$69.0\pm0.6$&$71.0\pm0.1$&$71.4\pm0.2$&$70.9\pm0.4$\\
    4-LAPGAN&$69.1\pm0.3$&$71.1\pm0.4$&$70.5\pm0.2$&$71.0\pm0.4$\\
    Deep CAE I&$68.0\pm0.4$&$70.1\pm0.3$&$70.9\pm0.4$&$70.2\pm0.1$\\
    Deep CAE II&$64.6\pm0.1$&$66.2\pm0.3$&$67.7\pm0.1$&$66.9\pm0.2$\\
    \hline
  \end{tabular}
\end{table}
\begin{figure}[!t]
\centering
\includegraphics[width=0.5\textwidth]{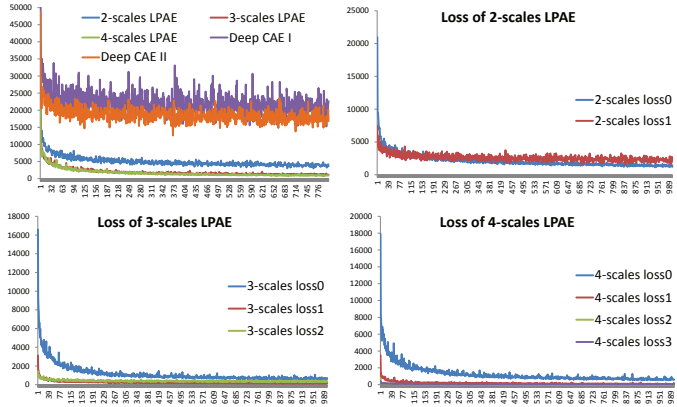}
\caption{Loss value against the number of training steps. ``2-scales loss0'' indicates the reconstruction loss of 2-scales LPAE at level 0.}
\end{figure}
\subsection{Convergence Analysis}

Figure 2 plots the loss value against the number of training steps, and the training procedures of all LPAE models were stopped after 30 epoches. As we can see, the converging speed was very fast for LPAE models, and the training procedure of LPAE was more stable than deep CAE. Besides, the loss of LPAE was much lower than deep CAE, which indicated that LPAE was more suitable for image generation task. The number of scales also had influence on the reconstruction loss. It's clear that the loss of 2-scales LPAE was larger than the rest of LPAE models. Thus it's important to chose appropriate number of scales regarding the image resolution. The key idea of this work is to break the learning procedure into successive refinements, which apparently worked well.

Adopting BN can help convergence, which is a well known trick in deep learning now. The results in Table 5 confirmed this view, but there was an exception that deep CAE II still converged when removing the BN layers. Figure 3 plots the performance of LPAE and deep CAE (with or without BN layers) against the number of epochs. As we can see, the performances of LPAE and deep CAE achieved the best very fast and stayed stable after 10 epochs. This result is helpful when applying LPAE in practice. BN has a very important influence on the performance of LPAE and deep CAE. The result at epoch 0 in Figure 3 is the performance of random filters. It's clear that LPAE and deep CAE perform worse than the random filters without BN layers, and using BN layers can lead to a drastic improvement of performance. Again deep CAE II showed a different behavior from other models after removing BN layers. As shown in Figure 3, we can see that its accuracy curve first went up and then went down after removing BN layers, which was different from other models. This phenomenon looks like over-fitting, but it's hard to believe that over-fitting problem would happen in unsupervised learning. It's the only difference between deep CAE II with other models that the architecture of deep CAE II is much deeper. Thus one possible explanation is that the deep architecture leads to more powerful learning capability, but also poses the problem of unstable gradients. Thus powerful learning capability made it avoid the training collapse, but unstable gradients leaded to unstable features. To confirm this view point needs more experiments in depth, and it's not the major concern of this paper. Thus we leave it to the future work.
\begin{table}[!t]
  \renewcommand{\arraystretch}{1.3}
  \caption{Convergence and BN. The symbol ``$\surd$'' indicates convergence, and the symbol ``$\times$'' indicates non-convergence.}
  \label{BN}
  \centering
  \begin{tabular}{lcc}
    \hline
    Algorithm&Adopt BN&Don't adopt BN\\
    \hline
    2-scales LPAE&$\surd$&$\times$\\
    3-scales LPAE&$\surd$&$\times$\\
    4-scales LPAE&$\surd$&$\times$\\
    2-scales LAPGAN&$\surd$&$\times$\\
    3-scales LAPGAN&$\surd$&$\times$\\
    4-scales LAPGAN&$\surd$&$\times$\\
    Deep CAE I&$\surd$&$\times$\\
    Deep CAE II&$\surd$&$\surd$\\
    \hline
  \end{tabular}
\end{table}
\begin{figure}[!t]
\centering
\includegraphics[width=0.5\textwidth]{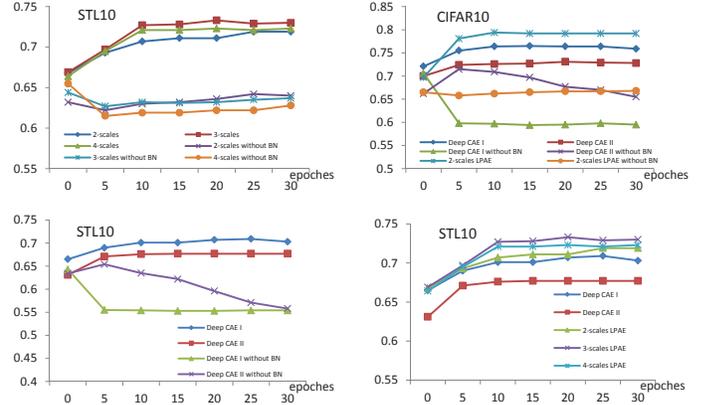}
\caption{Top left: LPAE with BN layers VS LPAE without BN layers on STL10; Top right: LPAE VS Deep CAEs on CIFAR10; Bottom left: Deep CAEs with BN layers VS Deep CAEs without BN layers on STL10; Bottom right: LPAEs VS Deep CAEs on STL10.}
\end{figure}

\section{Conclusion}
In this paper we embed deep auto-encoders into the framework of Laplacian pyramid, and apply the LPAE model to unsupervised representation learning. Experiments have shown some interesting results which benefit relative research and practical applications of deep auto-encoders approaches. First, scale-space representation like Laplacian pyramid benefitted the image representation learning. Second, for now the auto-encoder framework is more appropriate than generative adversarial nets to combine with Laplacian pyramid for unsupervised representation learning due to more stable training procedure. Third, the number of scales should be set appropriately regarding to the image resolution. Fourth, the learning procedure is efficient that the performances of the learned representations achieved the best very fast. Overall, the key idea of this work is to break the learning procedure into successive refinements, which aims at scale information learning and more stable training.

% or

\appendices

\ifCLASSOPTIONcompsoc
  \section*{Acknowledgments}
\else
  \section*{Acknowledgment}
\fi

Thanks for all the supports we have.

\ifCLASSOPTIONcaptionsoff
  \newpage
\fi



%\bibliographystyle{IEEEtran}
\bibliographystyle{IEEEtran}
\bibliography{IEEEabrv,ref}


%\newpage


%\vfill


\end{document}